%% file: main.tex
\documentclass{article}

\usepackage[final]{neurips_rl2019} % Uncomment for the camera-ready ``final'' version

\usepackage[utf8]{inputenc} % allow utf-8 input
\usepackage[T1]{fontenc}    % use 8-bit T1 fonts
\usepackage{hyperref}       % hyperlinks
\usepackage{url}            % simple URL typesetting
\usepackage{booktabs}       % professional-quality tables
\usepackage{amsfonts}       % blackboard math symbols
\usepackage{nicefrac}       % compact symbols for 1/2, etc.
\usepackage{microtype}      % microtypography
\usepackage{algorithm}
\usepackage{algorithmic}
\usepackage{mathtools}
\usepackage{graphicx}
\usepackage{cases}
\usepackage{array}
\usepackage{color}
\usepackage{float}
\usepackage{amssymb}
\usepackage{appendix}
\usepackage{wrapfig}
\usepackage{subfigure}
\usepackage{amsthm}
\usepackage{natbib}


\newcommand{\define}{\triangleq}
\newtheorem{theorem}{Theorem}

\newtheorem{definition}{Definition}
\newtheorem{assumption}{Assumption}

\title{$H_\infty$ Model-free Reinforcement Learning \\ with Robust Stability Guarantee}

\author{%
  Minghao Han \\
  Department of Control Science and Engineering\\
  Harbin Institute of Technology\\
  \texttt{mhhan@hit.edu.cn} \\
    \And
   {Yuan Tian}\\
   Department of Cognitive Robotics\\
   Delft University of Technology \\
   \texttt{y.tian-8@student.tudelft.nl} \\
     \And
     Lixian Zhang \\
   Department of Control Science and Engineering\\
   Harbin Institute of Technology \\
   \texttt{lixianzhang@hit.edu.cn} \\
  \And
   Jun Wang\\
   Department of Computer Science\\
   University College London \\
   \texttt{jun.wang@cs.ucl.ac.uk} \\
   \And
   {Wei Pan$^*$} \\
   Department of Cognitive Robotics\\
   Delft University of Technology \\
   \texttt{wei.pan@tudelft.nl} \\
}

\begin{document}
\maketitle
\let\thefootnote\relax\footnotetext{$*$ for correspondence.}

%===============================================================================

\begin{abstract}

Reinforcement learning is showing great potentials in robotics applications, including autonomous driving, robot manipulation and locomotion. However, with complex uncertainties in the real-world environment, it is difficult to guarantee the successful generalization and sim-to-real transfer of learned policies theoretically. In this paper, we introduce and extend the idea of robust stability and $H_\infty$ control to design policies with both stability and robustness guarantee. Specifically, a sample-based approach for analyzing the Lyapunov stability and performance robustness of a learning-based control system is proposed. Based on the theoretical results, a maximum entropy algorithm is developed for searching Lyapunov function and designing a policy with provable robust stability guarantee. Without any specific domain knowledge, our method can find a policy that is robust to various uncertainties and generalizes well to different test environments. In our experiments, we show that our method achieves better robustness to both large impulsive disturbances and parametric variations in the environment than the state-of-art results in both robust and generic RL, as well as classic control. Anonymous code is available to reproduce the experimental results at {\url{https://github.com/RobustStabilityGuaranteeRL/RobustStabilityGuaranteeRL}}.

\end{abstract}

%===============================================================================

\section{Introduction}

As a powerful learning control paradigm, reinforcement learning is extremely suitable for finding the optimal policy in tasks where the dynamics are either unknown or affected by severe uncertainty \citep{bucsoniu2018reinforcement}. Its combination with the deep neural network has boosted applications in autonomous driving \citep{sallab2017deep}, complicated robot locomotion \cite{hwangbo2019learning}, and skilful games like Atari \citep{mnih2015human} and Go \citep{silver2017mastering}. 
However, overparameterized policies are prone to become overfitted to the specific training environment, limiting its generalization to the various scenarios \cite{pinto2017robust}. Additionally, RL agents trained in simulation, though cheap to obtain, but likely suffer from the \textit{reality gap} problem \cite{koos2010crossing} when transferred from virtual to the real world. To overcome these drawbacks, various efforts are made to enhance the robustness of the policy \cite{jakobi1995noise, tobin2017domain,mordatch2015ensemble}, since a robust policy has a greater chance of successful generalization and transfer.

\textbf{Contribution} In this paper, we propose a unified framework of designing policies with both stability and robust performance guarantee against the various uncertainties in the environment. Without any specific domain knowledge, our method is able to find policy that is robust to large exogenous disturbances and generalizes well to different test environment. First, a novel model-free method for analyzing the Lyapunov stability and $H_\infty$ performance of the closed-loop system is developed. Based on the theoretical results, we propose the Robust Lyapunov-based Actor-Critic (RLAC) algorithm to simultaneously find the Lyapunov function and policy that can guarantee the robust stability of the closed-loop system. We evaluate RLAC on a simulated cartpole in the OpenAI gym \citep{brockman2016openai} environment and show that our approach is robust to: \textbf{i) Large impulsive disturbance:} The trained agent is able to recover when disturbed by adversary impulses 4-6 times of the maximum control input, while other baselines fail almost surely. \textbf{ii) Parametric Uncertainty:} The learned policy generalizes better than the baselines to different test environment settings (e.g. different mass and structural values).
%===============================================================================

\section{Preliminaries}
\label{sec:preliminaries}
\subsection{Markov Decision Process and Reinforcement Learning}
A Markov decision process (MDP) is a tuple, 
($S,A,c,P,\rho$), where $S$ is the set of states,
$A$ is the set of actions, $c (s,a)\in [0,\infty)$ is the cost function, $P (s'|s,a)$ is the transition probability function, and $\rho (s)$ is the starting state  distribution. $\pi(a|s)$ is a stationary policy denoting the probability of selecting action $a$ in state $s$. In addition, the cost function under stationary policy is defined as $c_\pi(s) \doteq \mathbb{E}_{a\sim\pi }c(s,a)$.

In this paper, we divide the state $s$ into two vectors, $s^1$ and $s^2$, where $s^1$ is composed of elements of $s$ that are aimed at tracking the reference signal $r$ while $s^2$ contains the rest. The cost function is defined as $\Vert s^1-r \Vert$, where $\Vert \cdot \Vert$ denotes the Euclidean norm.

\subsection{Robust Control Against Environment Uncertainty}

\begin{definition}
\label{def:mss}
The stochastic system is said to be stable in mean cost if $\lim_{t\rightarrow \infty }\mathbb{E}_{s_{t}} c_\pi(s_{t})=0$ holds for any initial condition $s_{0}\in \{s_{0}|c_\pi(s_{0})\leq b\}$. If $b$ is arbitrarily large then the stochastic system is globally stable in mean cost.
\end{definition}
To address the performance of the agent in the presence of uncertainty, the following definition is needed.
\begin{definition}
\label{def:l2}
The system is said to be mean square stable (MSS) with an $l_2$ gain less or equal than $\eta$, if the system is MSS when $w=0$, and the following holds for all $w\in l_2[0,+\infty)$,
\begin{equation}
\sum_{t=0}^\infty\mathbb{E}_{s_{t}}c_\pi(s_t) \leq \sum_{t=0}^\infty\mathbb{E}_{s_{t}}\eta^2 \Vert w(s_t)\Vert_2  \label{def: robust performance}
\end{equation}
where $\eta \in \mathbb{R}_+$. $ z(s_t)$ is the error output of the system and $w(s_t)$ is the uncertainty, which is composed of both environmental disturbance and modelling error.
\end{definition}
The robust performance guarantee \eqref{def: robust performance} holds for all $w$, is equivalent to guaranteeing the inequality for the worst case induced by $w$, i.e.,
\begin{equation}
\sup_{w} \sum_{t=0}^\infty\mathbb{E}_{s_{t}}c_\pi(s_t) -\eta^2 \Vert w(s_t)\Vert_2  \leq 0\label{eq: robust performance}
\end{equation}

\vspace{-0.3cm}
\section{Main Results} \label{sec:main result}
\vspace{-0.2cm}

\subsection{Lyapunov-based $H_\infty$ Learning Control}
In this section, we propose the main assumptions and a new theorem.

\begin{assumption}\label{stationary assumption}
The stationary distribution of state $q_\pi(s)\define\lim_{t\rightarrow\infty}P(s|\rho,\pi,t)$ exists.
\end{assumption}

\begin{assumption}\label{initial state assumption}
There exists a positive constant $b$ such that $\rho(s)> 0, \forall s\in\{s|c_\pi(s)\leq b\}$.
\end{assumption}

The core theoretical results on analyzing the stability and robust performance of the closed-loop system with the help of Lyapunov function and sampled data are presented. 
The Lyapunov function is a class of continuously differentiable semi-positive definite functions $L : \mathcal{S} \to \mathbb{R}_+$. The general idea of exploiting Lyapunov function is to ensure that the derivative of Lyapunov function along the state trajectory is semi-negative definite so that the state goes in the direction of decreasing the value of Lyapunov function and eventually converges to the set or point where the value is zero.

\begin{theorem}\label{them:robust mss}
If there exists a continuous differentiable function $L:\mathcal{S}\rightarrow
%TCIMACRO{\U{211d} }%
%BeginExpansion
\mathbb{R}
%EndExpansion
_{+}$\ and positive constants $\alpha _{1}$, $\alpha _{2}$, $\alpha_{3},\eta, k_1, k_2$ such that
\begin{gather}
\alpha _{1}c_\pi\left( s\right) \leq L(s)\leq \alpha _{2}c_\pi\left( s\right)
\label{robust stability-1}\\
\mathbb{E}_{\beta(s)}(\mathbb{E}_{s^{\prime }\sim P_{\pi }}L(s^{\prime
})-L(s))<  \mathbb{E}_{\beta(s) }[\eta \Vert w(s) \Vert -(\alpha_{3}+1)c_\pi\left( s\right)]
\label{robust stability-2}
\end{gather}%
holds for all $\{\rho|\mathbb{E}_{s_0\sim\rho}c_\pi(s_0)\leq k_1\}$ and $\{w|\Vert w\Vert\leq k_2\}$. $\beta_\pi(s)\define \lim_{N\rightarrow\infty}\frac{1}{N}\sum_{t=0}^N P(s_t=s|\rho,\pi,t)$ is the sampling distribution. 
Then the system is mean square stable and has $l_2$ gain no greater than $\eta/(\alpha_3 + 1)$. If the above holds for $\forall k_1, k_2\in \mathbb{R}_+$, then the system is globally mean square stable with finite $l_2$ gain. 
\end{theorem}
Proof of Theorem~\ref{them:robust mss} is given in Appendix~\ref{proof of robust mss}.

\subsection{Learning the Adversarial Disturber}

In our setting, in addition to the control policy $\pi$, a disturber policy $\mu(w|s)$ is introduced to actively select the worst disturbance for a given state. More specifically, the adversarial disturber seeks to find the disturbance input $w$ over which the system has the greatest $l_2$ gain, i.e. maximizing the following cost function,
{\small
\begin{equation}
    \max_{\theta_\mu} J(\mu) = \mathbb{E}_{\beta(s), \mu(w|s)}(c_\pi(s)-\eta^2 \Vert w \Vert)
\end{equation}
where $\theta_\mu$ is the parameter of the disturber policy $\mu$.
}
\section{Algorithm}\label{sec:algorithm}
In this section, based on the theoretical results in Section~\ref{sec:main result}, we propose an actor-critic style algorithm with robust stability guarantee (RLAC).

In this algorithm, we include a critic Lyapunov function $L_c$ to provide the policy gradient, which satisfies $L(s) = \mathbb{E}_{a\sim \pi} L_c(s,a)$. Through Lagrangian method, the objective function for $\pi$ is obtained as follow,
{\small
\begin{equation}
\begin{aligned}
J(\pi) &= \mathbb{E}_{(s,a,w,c,s')\sim\mathcal{D}}\left[ \beta \log(\pi(f_{\theta_\pi}(\epsilon,s)|s))+ \lambda \Delta L(s,a,w,c,s')\right]\\
\Delta L(s,a,w,c,s')&=\left(L_c(s',f_{\theta_\pi}(\epsilon,s'))-L_c(s,a)+(\alpha_3+1)c - \eta^2\Vert w \Vert_2\right)
\label{RLAC}
\end{aligned}
\end{equation}
}
where $\pi$ is parameterized by a neural network $f_{\theta_\pi}$ and $\epsilon$ is an input vector consisted of Gaussian noise. In the above objective, $\nu$ and $\lambda$ are the positive Lagrangian multipliers, of which the values are adjusted automatically. The gradient of \eqref{RLAC} with respect to the policy parameter $\theta_\pi$ is approximated by {\small
\begin{equation}
    \nabla_{\theta} J(\pi) =\mathbb{E}_{\mathcal{D}}\left[ \nabla_\theta \nu \log(\pi_\theta(a|s)) + \nabla_a \nu \log(\pi_\theta(a|s))\nabla_\theta f_\theta(\epsilon,s) + \lambda\nabla_{a'}L_c(s',a')\nabla_\theta f_\theta(\epsilon,s')\right]
    \label{algorithm:RLAC_policy_gradient}
\end{equation}
}

The Lyapunov function is updated through minimizing the following objective
{\small
\begin{gather}
    J(L_c) = \mathbb{E}_{(s,a)\sim \mathcal{D}}\left[\frac{1}{2}(L_c(s,a)-L_{\text{target}}(s,a))^2\right]
\end{gather}
}
We use the sum of cost over a finite time horizon $N$ as the Lyapunov candidate, i.e.
{\small
\begin{gather}
    L_{\text{target}}(s,a) = \sum_{t=t_0}^N c(s_t,a_t)
\end{gather}
}
which has long been exploited as the Lyapunov function in establishing the stability criteria for model predictive control (MPC) \citep{mayne2000constrained}. The pseudo-code of RLAC is presented in Algorithm~\ref{algo:RLAC}.

\section{Experimental Results}
\label{sec:experiment result}
In this section, we evaluate the robustness of RLAC against i) large impulsive disturbances; ii) parametric uncertainty. Setup of the experiment is referred to Appendix~\ref{Experiment setup}.
%iii) different parameter initializations. 

\subsection{Robustness to Impulsive Disturbances}

\vspace{-0.3cm}
\begin{figure}[htb]
    \centering
    
    \subfigure[Visualization of the disturbance]{
    \includegraphics[scale=0.32]{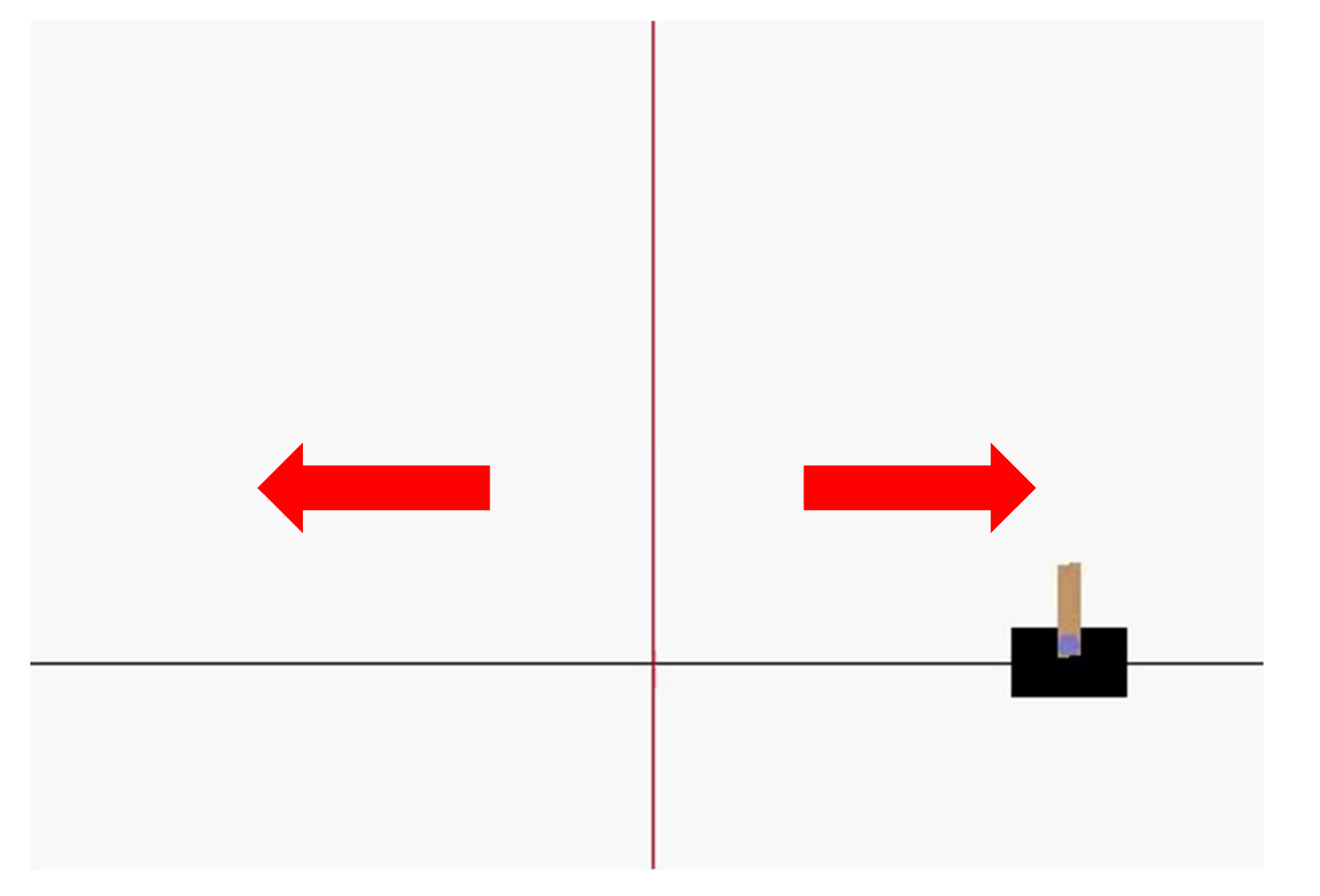}
    }
    \subfigure[Death rate of the cartpole]{
    \includegraphics[scale = 0.27]{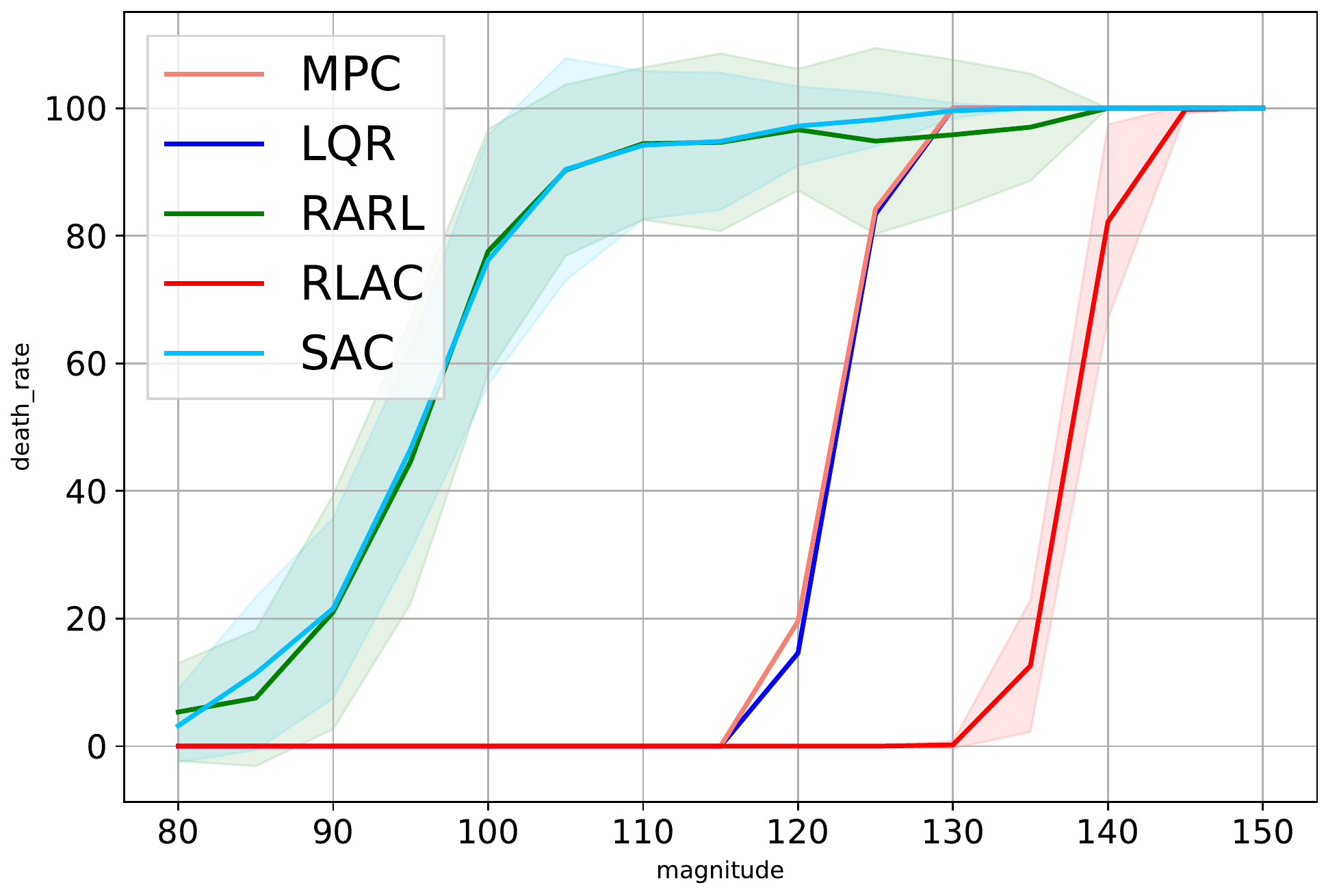}
    }
    \caption{(a) Direction of the disturbance applied on the cartpole, which is dependent on the relative position of cart concerning the origin. (b) The death rate of agents trained by RLAC, RARL, SAC, MPC  and LQR in the presence of impulsive force $F$ with different magnitudes. The trained policies are initialized by 10 random seeds. The policies with different initializations are evaluated equally for 500 episodes. The line indicates the average death rate of these policies and the shadowed region shows the 1-SD confidence interval.}
    \label{fig:impulsive disturbance}
\end{figure}
We evaluate the robustness of the agents trained by RLAC and baselines against unseen exogenous disturbance.
We measure the robust performance via the death rate, i.e., the probability of pole falling after impulsive disturbance. 
As observed in the figure, RLAC gives the most robust policy against the impulsive force. It maintains the lowest death rate throughout the experiment, far more superior than SAC and RARL. Moreover, RLAC performs even better than MPC and LQR, which possess the full information of the model and are available.

\subsection{Robustness to Parametric Uncertainty}

In this experiment, we evaluate the trained policies in environments with different parameter settings. In the training environment, the parameter \textit{length of pole} $l=0.5$ and \textit{mass of cart} $m_c=1$, while during evaluation $l$ and $m_c$ are selected in a 2-D grid with $l\in[0.2,2.0]$ and $m_c\in[0.4,2.0]$. 

\begin{figure}[h]
    \centering
    \includegraphics[scale = 0.47]{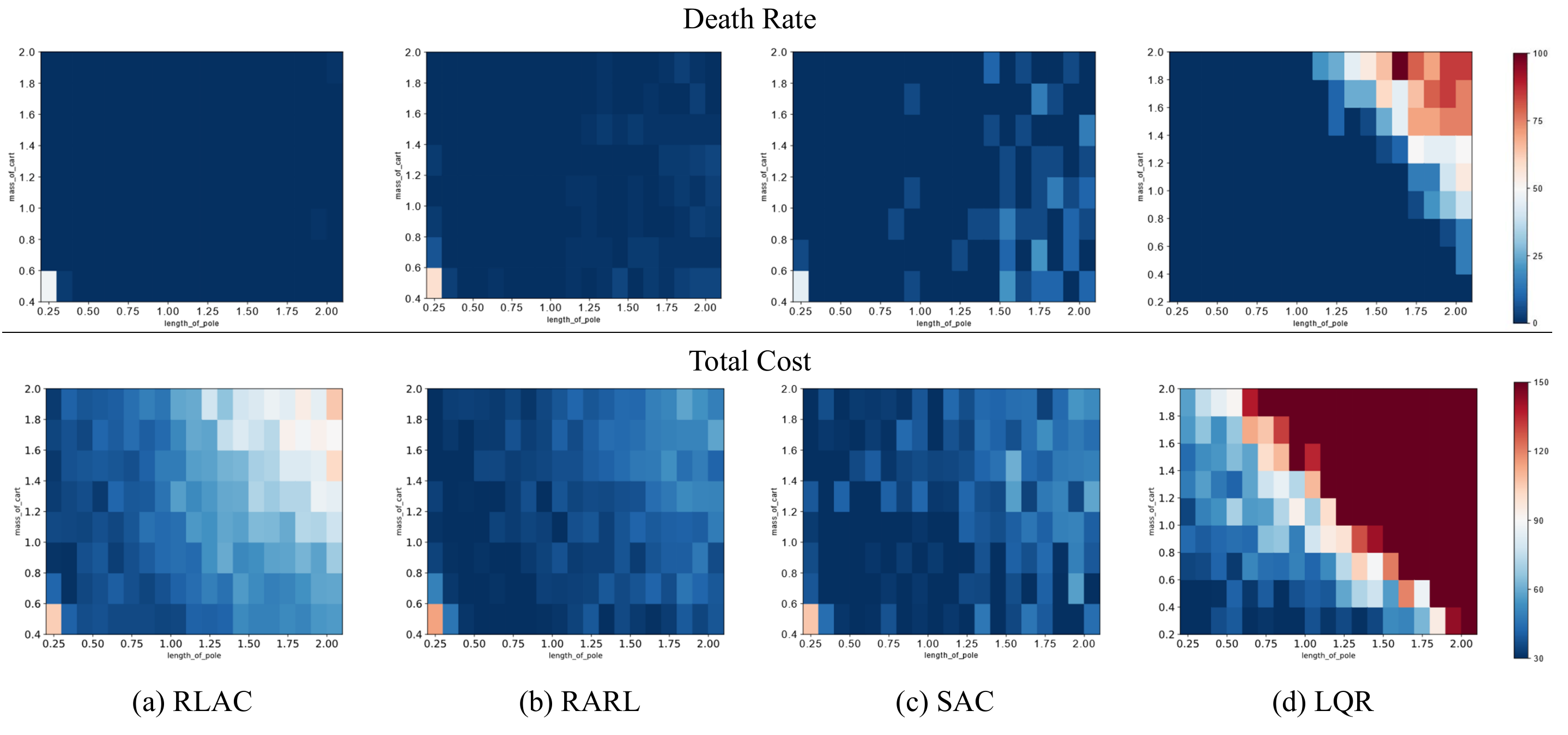}
    \caption{Death rate and total costs of agents trained by RLAC, RARL, SAC and LQR in the presence of different parametric uncertainty which are \emph{unseen} during training and different from dynamic randomization. $l$ (X-axis) and $m_c$ (Y-axis) vary with the step size of $0.1$ and $0.2$ respectively. At each point of the parameter grid, the results are averaged between the agents with different initializations over 100 episodes.
    }
    \label{fig:parametric uncertainty}
\end{figure}

As shown in the heat maps in \autoref{fig:parametric uncertainty}, RLAC achieves the lowest death rate (zero for the majority of the parameter settings) and obtains reasonable total cost (lower than 100). The total cost of RLAC is slightly higher than SAC and RARL since the agents hardly die and sustain longer episodes. Compared to SAC, RARL achieves lower death rate and comparable total cost performance. LQR performs well in the region where parameters are close to the nominal model but deteriorates soon as parameters vary. All of the model-free methods outperform LQR in terms of robustness to parametric uncertainty, except for the case of low $l$ and $m_c$ (left bottom of the grid). This is potentially due to the overparameterized policy does not generalize well to the model where dynamic is more sensitive to input than the one used for training.

%===============================================================================
\vspace{-0.2cm}

%\bibliography{ref}  % .bib
\bibliographystyle{unsrtnat}

\newpage
\appendix

\section{Proof of \autoref{them:robust mss}} \label{proof of robust mss}
\begin{proof}
The existence of sampling distribution $\mu_\pi(s)$ is guaranteed by the existence of $q_\pi(s)$ (Assumption~\ref{stationary assumption}). Since the sequence $\{P(s|\rho,\pi,t), t\in\mathbb{Z}_+\}$ converges to $q_\pi(s)$ as $t$ approaches $\infty$, then by the Abelian theorem, the sequence $\{\frac{1}{N}\sum_{t=0}^N P(s|\rho,\pi,t), N\in\mathbb{Z}_+\}$ also converges and $\mu_\pi(s) = q_\pi(s)$.
    Combined with the form of $\mu_\pi$, Eq.(\ref{robust stability-2}) infers that
    \begin{equation}
    \begin{aligned}
        &\int_\mathcal{S}\lim_{N\rightarrow\infty}\frac{1}{N}\sum_{t=0}^N P(s|\rho,\pi,t)(\mathbb{E}_ {P_{\pi}(s^{\prime}|s)}L(s^{\prime})-L(s))\mathrm{d}s
        < \mathbb{E}_{\beta(s) }[\eta \Vert w(s) \Vert -(\alpha_{3}+1)c_\pi\left( s\right)]
    \label{proof:theorem MSS 1}
    \end{aligned}
    \end{equation}
    
    First, the stability of the system in mean cost will be proved. 
    According to Eq.(\ref{robust stability-1}), $L(s)\leq \alpha_2 c_\pi(s)$ for all $s\in\mathcal{S}$ and consider that $P(s|\rho,\pi,t)\leq 1$,
    \begin{equation*}
    P(s|\rho,\pi,t)L(s)\leq \alpha_2 c_\pi\left(s\right), \forall s\in \mathcal{S}, \forall t\in \mathbb{Z}_+
    \end{equation*}
    On the other hand, the sequence $\{\frac{1}{N}\sum_{t=0}^N P(s|\rho,\pi,t)L(s), N\in\mathbb{Z}_+\}$ converges pointwise to the function $q_\pi(s)L(s)$.
    According to the Lebesgue's Dominated convergence theorem\citep{royden1968real}, if a sequence ${f_n(s)}$  converges pointwise to a function $f$ and is dominated by some integrable function $g$ in the sense that,
    \begin{equation*}
    \vert f_n(s) \vert \leq g(s), \forall s\in \mathcal{S},\forall n
    \end{equation*}
    Then 
    \begin{equation*}
     \lim_{n\rightarrow\infty}\int_\mathcal{S}f_n(s)\mathrm{d}s= \int_\mathcal{S}\lim_{n\rightarrow\infty}f_n(s)\mathrm{d}s
    \end{equation*}
    Thus the left hand side of Eq.(\ref{proof:theorem MSS 1})
    \begin{equation*}
    \begin{aligned}
     &\int_\mathcal{S}\lim_{N\rightarrow\infty}\frac{1}{N}\sum_{t=0}^N P(s|\rho,\pi,t)(\int_\mathcal{S} P_{\pi}(s^{\prime}|s)L(s^{\prime})\mathrm{d}s^{\prime}-L(s))\mathrm{d}s\\
     =&\lim_{N\rightarrow\infty}\frac{1}{N}(\sum_{t=1}^{N+1} \mathbb{E}_{P(s|\rho,\pi,t)}L(s)-\sum_{t=0}^{N} \mathbb{E}_{P(s|\rho,\pi,t)}L(s))\\
     =&\lim_{N\rightarrow\infty}\frac{1}{N} \left( \mathbb{E}_{ P(s|\rho,\pi,N+1)}L(s)-\mathbb{E}_{\rho(s)}L(s)\right)
    \end{aligned}
    \end{equation*}
    First, when $w(s)\equiv 0$, Eq.(\ref{proof:theorem MSS 1}) infers
    \begin{equation}
    \lim_{N\rightarrow\infty}\frac{1}{N} \left( \mathbb{E}_{P(s|\rho,\pi,N+1)}L(s)-\mathbb{E}_{ \rho(s)}L(s)\right)
    <-(\alpha_3+1)\lim_{t\rightarrow\infty}\mathbb{E}_{P(s|\rho,\pi,t)}c_\pi\left(s\right)
    \end{equation}
    Since $\mathbb{E}_{\rho(s)}L(s)$ is a finite value and $L$ is semi-positive definite, it follows that
    \begin{equation}
    \lim_{t\rightarrow\infty}\mathbb{E}_{ P(s|\rho,\pi,t)}c_\pi\left(s\right)\leq \lim_{N\rightarrow\infty}\frac{1}{N} (\frac{1}{\alpha_3+1}\mathbb{E}_{\rho(s)}L(s))
    = 0 \label{final eq}
    \end{equation}
    Suppose that there exists a state $s_0\in\{s_0|c_\pi(s_0)\leq b\}$ such 
    $\lim_{t\rightarrow\infty}\mathbb{E}_{ P(s|s_0,\pi,t)}c_\pi\left(s\right)=c$, $c>0$ or $\lim_{t\rightarrow\infty}\mathbb{E}_{ P(s|s_0,\pi,t)}c_\pi\left(s\right)=\infty$. Consider that $\rho(s_0)> 0$ for all starting states in $\{s_0|c_\pi(s_0)\leq b\}$ (Assumption~\ref{initial state assumption}), then $\lim_{t\rightarrow\infty}\mathbb{E}_{s_t\sim P(\cdot|\pi,\rho)}c_\pi\left(s_t\right)>0$, which is contradictory with Eq.(\ref{final eq}). Thus $\forall s_0\in \{s_0|c_\pi(s_0)\leq b\}$, $\lim_{t\rightarrow\infty}\mathbb{E}_{ P(s|s_0,\pi,t)}c_\pi\left(s\right)=0$.
    Thus the system is stable in mean cost by Definition~\ref{def:mss}.

Next, the $H_{\infty}$ performance of the system will be proved. When $w(s)\neq 0$, Eq.(\ref{proof:theorem MSS 1}) infers
\begin{equation}
    \lim_{N\rightarrow\infty}\frac{1}{N} \left( \mathbb{E}_{P(s|\rho,\pi,N+1)}L(s)-\mathbb{E}_{ \rho(s)}L(s)\right)
    <\mathbb{E}_{\beta(s) }[\eta \Vert w(s) \Vert -(\alpha_{3}+1)c_\pi\left( s\right)]
    \end{equation}
    Since $\mathbb{E}_{ \rho(s)}L(s)$ is a finite and 
Due to the semi-definiteness of $c_\pi$ and $\mathbb{E}_{P(s|\rho,\pi,N+1)}L(s)$ is semi-positive definite, one has
\begin{equation}
\begin{aligned}
    % \mathbb{E}_{\beta(s) }[(\alpha_{3}+1)c_\pi\left( s\right)-\eta \Vert w(s) \Vert ]&<0\\
    \lim_{N\rightarrow\infty}\frac{1}{N}\sum_{t=0}^{N} \mathbb{E}_{P(s|\rho,\pi,t)}\left[\left(\alpha_{3}+1\right)c_\pi\left( s\right)-\eta \Vert w(s) \Vert \right]&<0
\end{aligned}
\end{equation}
The above inequality only holds if $\sum_{t=0}^{\infty} \mathbb{E}_{P(s|\rho,\pi,t)}\left[\left(\alpha_{3}+1\right)c_\pi\left( s\right)-\eta \Vert w(s) \Vert \right]<0$, thus the system has $l_2$ gain less than $\frac{\eta}{\alpha_{3}+1}$, which concludes the proof.
\end{proof}

\section{Algorithm}\label{Algorithm Details}

\begin{algorithm}[H]
   \caption{Robust Lyapunov-based Actor Critic (RLAC)}
   \label{algo:RLAC}
\begin{algorithmic}
   \STATE Initialize replay buffer $R$ and the Lagrangian multiplier $\lambda$, $\beta$, learning rate $\alpha$
   \STATE Randomly initialize Lyapunov critic network $L_c(s, a)$, actor $\pi(a|s)$, disturber $\mu(d|s)$ with parameters $\phi_{L_c}$, $\theta_\pi$, $\theta_\mu$
    \STATE Initialize the parameters of target networks with $\overline{\phi}_{L_c}\leftarrow\phi_{L_c}$, $\overline{\theta}_\pi\leftarrow\theta_\pi$
   \FOR{each iteration}
   \STATE Sample $s_0$ according to $\rho$
   \FOR{$t$ \text{in} $N_c$}
   \STATE Sample $a_t$ from $\pi_a(s)$ and $d_t$ from $\pi_d(s)$ and step forward
   \STATE Observe $s_{t+1}$, $c_{t}$ and store $(s_t,a_t, d_t,c_t,s_{t+1})$ in $R$
   \ENDFOR
   \FOR{$i$ \text{in} $N_u$}
   \STATE Sample minibatches of $N$ transitions from $R$
   \STATE Estimate policy gradient:
   \begin{equation}
    \begin{aligned}
    \phi_{L_c} & \leftarrow \phi_{L_c}  +\alpha\nabla_{\phi_{L_c}} J(L_c)\notag\\
    \theta_\pi &\leftarrow \theta_\pi +\alpha \nabla_{\theta_\pi} J(\pi) \notag\\
    \theta_\mu &\leftarrow \textit{PolicyOptimizer}( J(\pi_\mu))
    \notag
    \end{aligned}
    \end{equation}
   \STATE Update Lagrangian multipliers $\lambda$, $\beta$ and the parameter of target networks, $\overline{\phi}_{L_c}$,  $\overline{\theta}_\pi$
   \ENDFOR
   \ENDFOR
   
\end{algorithmic}
\end{algorithm}
\hspace{-0.2cm}

\section{Experiment Setup}\label{Experiment setup}

\begin{figure}[htb]
    \centering
    \includegraphics[scale=0.35]{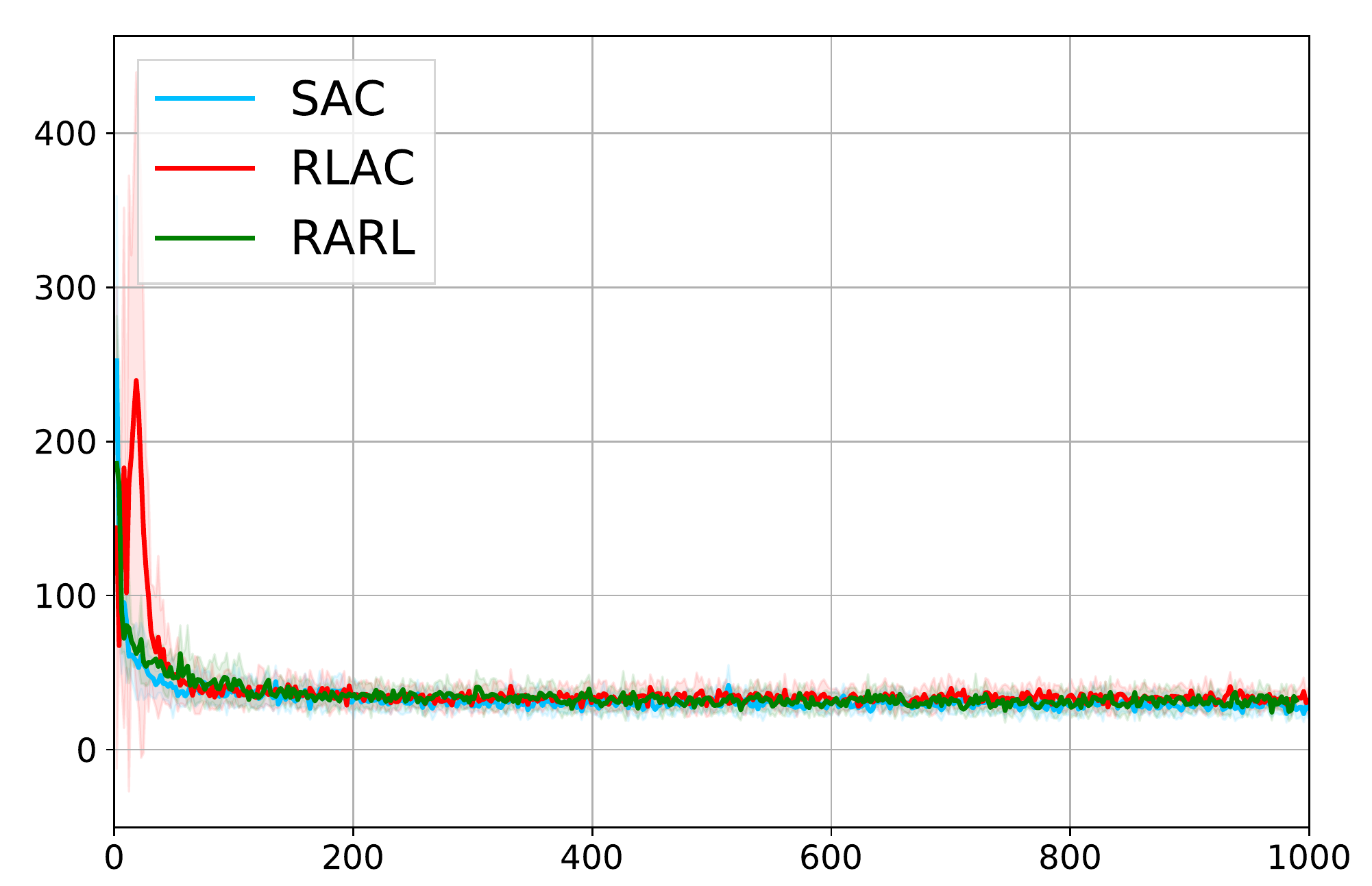}
    
    \caption{Cumulative cost curves for RLAC, RARL and SAC; the shadowed region shows the 1-SD confidence interval over 10 random seeds and the X-axis indicates the total time steps in thousand.}

    \label{fig:training curve}
\end{figure}
We compare with robust adversarial reinforcement learning (RARL, \citep{pinto2017robust}) which is considered to be the model-free robust RL baseline. RARL demonstrated great robustness to both exogenous disturbance and parametric disturbance on a series of continuous control tasks. In the implementation of RARL \cite{pinto2017robust}, TRPO was used as the policy optimizer, though other policy optimization methods are also comparable with the framework. Without loss of generality, we use soft actor-critic (SAC) \citep{haarnoja2018soft} as the policy optimizer for RARL, given that both sample efficiency and final performance of SAC exceeds the state of art on a series of continuous control benchmarks. We also include SAC as the generic RL baseline. Linear Quadratic Regulator (LQR) and MPC are included as the model-based baseline, which is guaranteed to find the optimal analytic solution given the well established linear model and weighting matrices. Hyperparameters and other details for the implementation of RLAC are referred to Appendix~\ref{Hyperparameter}.

We evaluate RLAC and baselines on a simulated cartpole in the OpenAI gym environment (all the units are dimensionless), details of the environment is referred to Appendix~\ref{Experiment setup}. The agent is expected to sustain the pole vertically at $x=0$. Both the RLAC and baseline agents are trained under the static environment setting, i.e., with unchanged model parameters and dynamics, and evaluated in the environment variants with unseen uncertainty.
During training, both RLAC and RARL generate disturbance input to affect the performance of the policy. The cost and state at the next step are determined by the action and disturbance jointly. The magnitude of the disturbances is held no larger than $5$, while the actions are below $20$.

Each algorithm is trained for the same amount of global time steps over $10$ random seeds with optimized hyperparameters. The total cost of rollouts during training is shown in \autoref{fig:training curve}. As shown in the figure, all the algorithms converge eventually, achieving similar final performance in terms of return.

In the cartpole experiment, the agent is to sustain the pole vertically at the central position. This is a modified version of cartpole in \citet{brockman2016openai} with continuous action space. The action is the horizontal force on the cart($F\in[-20, 20]$). 
 The position $x$ is limited to $(-x_{\text{threshold}}, x_{\text{threshold}})$, $x_{\text{threshold}}=10$. $\theta$ is the angle position with respect to the vertical direction, $\theta \in (-\theta_{\text{threshold}},\theta_{\text{threshold}})$ where $\theta_{\text{threshold}}=0.349 \textit{ rad}$.
The cost function $ c= (\frac{x}{x_{\text{threshold}}})^2 + 20(\frac{ \theta}{\theta_{\text{threshold}}})^2$. The agent is initialized randomly between $x\in[-5,5]$ while other variables initialized in $[-0.2, 0.2]$.
The maximum length of episodes is 250.

\subsection{Robustness to Impulsive Disturbances}
We evaluate the robustness of the agents trained by RLAC and baselines against unseen exogenous disturbance. To show this, we implement an impulsive adversarial force on the cartpole, ranging from 80 to 120 which is far more larger than the maximum action input and disturbance during training. The impulsive disturbance is analogous to a sudden hit or strong wind applied on the cart. The disturbance acts in the direction of pushing the cart away from the origin, as shown in \autoref{fig:impulsive disturbance} (a), and only takes place at the $100$\textit{th} time step.

We use the death rate instead of total cost because the failure will end the episode in advance and result in a rather low total cost. Under different impulse magnitudes, the policies trained by RLAC and baselines over different initializations are evaluated for 500 times, and results are shown in \ref{fig:impulsive disturbance} (b).

\subsection{Robustness to Parametric Uncertainty}

The trained agents with different initializations are evaluated with an equal number of episodes, and at each point of the parameter grid, the agents are evaluated for 100 times. For the same reason discussed in the previous subsection, total cost together with death rate are shown in \autoref{fig:parametric uncertainty} to demonstrate the robustness of RLAC and baselines.

\section{Hyperparameters}\label{Hyperparameter}

\begin{table}[htb] 
\begin{center}
\begin{tabular}{l|c}
\hline
Hyperparameters&Value\\\hline
Minibatch size& 256\\
Actor learning rate & 1e-4\\
Lyapunov learning rate & 3e-4\\
Time horizon $N$ & 10\\
$N_c$& 150\\
$N_u$& 50\\
Target entropy& -1\\
Soft replacement($\tau$) &0.005\\
Discount($\gamma$) & 0.995\\
$\eta$&1\\
$\alpha_3$&1\\\hline
\end{tabular}
\end{center}
\caption{RLAC Hyperparameters}
\end{table}
 For the policy network, we use a fully-connected MLP with two hidden layers of 64 units, and ReLU nonlinearities, outputting the mean and standard deviations of a Gaussian distribution. For the Lyapunov network, we use a fully-connected MLP with two hidden layers of 64 units, and ReLU nonlinearity respectively, outputting the Lyapunov value. We adopt the same invertible squashing function technique as \citet{haarnoja2018soft} to the output layer of policy network.

\end{document}